\documentclass[3p,times]{elsarticle}

\usepackage{ecrc}
\usepackage{graphicx}
\usepackage{array}
\usepackage{multirow}
\usepackage{CJK}
\usepackage{graphicx}
\usepackage{algpseudocode}
\usepackage{amsmath}
\usepackage{CJK}
\usepackage{amssymb}
\usepackage{subfigure}
\usepackage{subfig}
\usepackage{caption}
\usepackage[ruled]{algorithm2e}
\usepackage{booktabs}
\usepackage{balance}
\usepackage{enumitem}
\usepackage{float}

\volume{00}

\firstpage{1}

\runauth{}

\jid{TCS}

\usepackage{amssymb}

\usepackage{lineno}

\begin{document}
	
	\begin{frontmatter}

		\dochead{}

		\title{Dual-path CNN with Max Gated block for Text-Based Person Re-identification}

		\author[label1]{Tinghuai Ma \corref{cor1}}
		\author[label1]{Mingming Yang}
		\author[label2]{ Huan Rong}
		\author[label3]{Yurong Qian}
		\author[label4]{Yuan Tian}
		\author[label5]{NajlaAl-Nabhan}

		
		\address[label1]{School of Computer \& Software, Nanjing University of
			information science and Technology, Jiangsu, Nanjing 210-044, China}
		\address[label2]{School of Artificial Intelligence, Nanjing University of
			information science and Technology, Jiangsu, Nanjing 210-044, China }
		\address[label3]{Xinjiang University, Urumqi, 830-008, China}
		\address[label4]{Nanjing Institute of Technology, Jiangsu, Nanjing 211-167, China }
		\address[label5]{Dept. Computer Science, KingSaud University, Riyadh 11362, Saudi Arabia. }
		%
		
		\cortext[cor1]{Corresponding author: thma@nuist.edu.cn}

		\begin{abstract}
			Text-based person re-identification(Re-id) is an important task in video surveillance, which consists of retrieving the corresponding person's image given a textual description from a large gallery of images. It is difficult to directly match visual contents with the textual descriptions due to the modality heterogeneity. On the one hand, the textual embeddings are not discriminative enough, which originates from the high abstraction of the textual descriptions. One the other hand, 
			Global average pooling (GAP) is commonly utilized to extract more general or smoothed features implicitly but ignores salient local features, which are more important for the cross-modal matching problem. With that in mind, a novel Dual-path CNN with Max Gated block (DCMG) is proposed to extract discriminative word embeddings and make visual-textual association concern more on remarkable features of both modalities. The proposed framework is based on two deep residual CNNs jointly optimized with cross-modal projection matching (CMPM) loss and cross-modal projection classification (CMPC) loss to embed the two modalities into a joint feature space. First, the pre-trained language model, BERT, is combined with the convolutional neural network (CNN) to learn better word embeddings in the text-to-image matching domain. Second, the global Max pooling (GMP) layer is applied to make the visual-textual features focus more on the salient part. To further alleviate the noise of the maxed-pooled features, the gated block (GB) is proposed to produce an attention map that focuses on meaningful features of both modalities. Finally, extensive experiments are conducted on the benchmark dataset, CUHK-PEDES, in which our approach achieves the rank-1 score of 55.81\% and outperforms the state-of-the-art method by 1.3\%.
			
		\end{abstract}
		
	\end{frontmatter}

	%

	
		\begin{figure*}
		\begin{center}
			\includegraphics[width=1.0\linewidth]{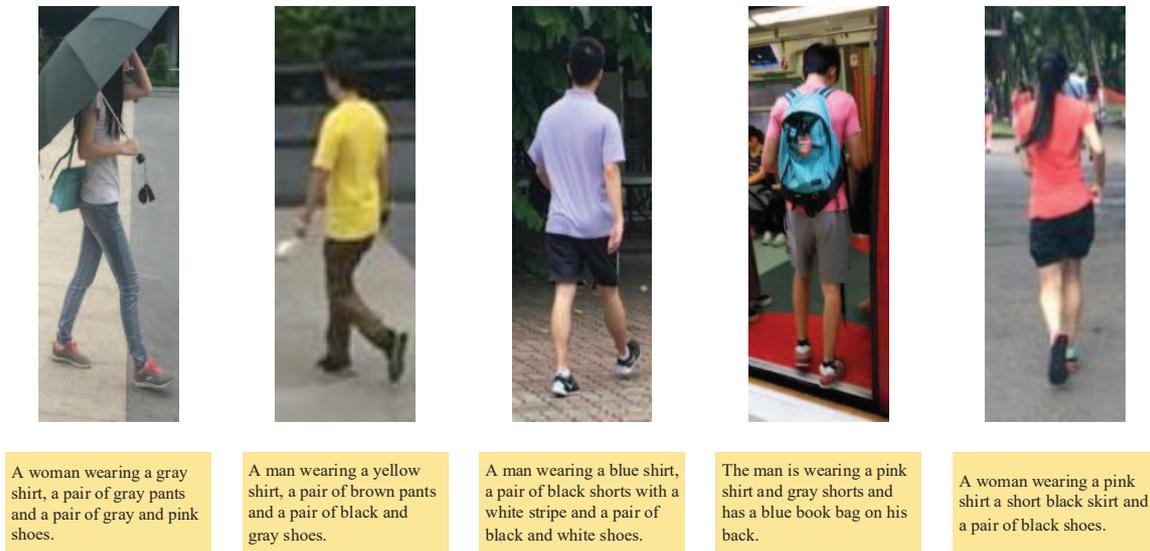} \ \\
		\end{center}
		\caption{Example sentence descriptions from dataset, CUHK-PEDES, that describe persons' appearances in detail. }
		\label{fig:dataset}
	\end{figure*}

	\section{Introduction}
	With the increase in public safety demands and video surveillance \cite{DBLP:journals/tip/QiaoWSC20,DBLP:journals/tmm/LokocBSMA18,DBLP:journals/tmm/Kordopatis-Zilos19}, manual person search in large-scale videos is unrealistic. Person re-identification (Re-id) \cite{DBLP:journals/tip/LinWYXY20,DBLP:journals/tip/FengLX20,DBLP:journals/tip/LiZH20,DBLP:journals/tip/TangYWSG20} is an important task for fast and accurate retrieval in big visual data. The methods of Re-id are mainly classified into three categories according to the query type, e.g., image-based query \cite{DBLP:journals/pr/RenXL19}, attribute-based query \cite{DBLP:journals/pr/LiYH20}, and text-based query \cite{DBLP:conf/iccv/Sarafianos0K19}. However, image-based and attribute-based Re-id have major limitations and might not be suitable for practical application. First, image-based person Re-id requires at least one photo of the queried person, which is very difficult to obtain in practice. Second, attribute-based person Re-id has limited capability of describing the appearance of persons. In contrast, text-based person Re-id can precisely describe the details of persons of interest to overcome such limitations. As shown in Figure \ref{fig:dataset}, adequate and comprehensive information can be provided including appearance, clothes, and action via language descriptions. Therefore, we study the task of text-based person Re-id in this paper.
	
	Specifically, text-based person Re-id aims to retrieve the corresponding person image according to a textual description from a large gallery of images. It is an extremely challenging task to directly match visual contents with the textual descriptions. Its challenges originate from modality heterogeneity between the high abstraction of the textual descriptions and the intuitive expression of images. To address the cross-modal matching problem, most existing methods rely on a relatively similar procedure to obtain most discriminative visual-textual feature embeddings and the most reasonable matching metric: (i) extract discriminative visual embeddings using a deep convolutional neural network (CNN) \cite{DBLP:journals/pr/LiMJLSZ20,DBLP:journals/pr/WongL20,DBLP:journals/ijon/LiaoWYZM20}, (ii) extract textual embeddings using a bi-LSTM or CNN, and (iii) propose a loss function that measures the distance between the two embeddings, such as cross-modal projection matching (CMPM) loss and cross-modal projection classification (CMPC) loss \cite{DBLP:conf/eccv/ZhangL18}, which increase the variance between unmatched samples and the association between the matched ones.
	
	Although the prior work \cite{DBLP:conf/wacv/ChenXL18,DBLP:conf/aaai/JingSW00T20,DBLP:conf/icassp/WangBWWQL19} has continuously improved the accurancy of text-based person Re-id, the performance remains unsatisfactory. On the one hand, the textual embeddings are not discriminative enough. The inherently sequential nature of bi-LSTM restricts parallelization within training examples, which limits the length of the text descriptions and destroys the integrity of the text description. CNN can overcome the limitation of parallelization, but it ignores the sequence of textual descriptions. On the other hand, visual-textual association concerns more on remarkable features of both modalities for more reasonable similarity measurement, but the traditional methods extract more general or smoothed features implicitly and ignore salient local features via global average pooling (GAP).
	
	In this paper, we propose a novel Dual-path CNN with Max Gated block (DCMG) to extract discriminative word embeddings and make visual-textual association concern more on remarkable features of both modalities. The proposed framework is based on two deep residual CNNs jointly optimized with CMPC loss and CMPM loss to embed the two modalities into a joint feature space. First, we apply Bidirectional Encoder Representations from Transformers (BERT) \cite{DBLP:conf/naacl/DevlinCLT19}, Where the order of the sequence can be utilized by the positional encoding, to learn the contextual relations between the words in a textual description. Our experiments demonstrate that BERT can be successfully combined with CNN to learn discriminative word embeddings and significantly improve the performance of the model. Second, to make the visual-textual features focus more on discriminative characteristics, i.e. salient part, the global max pooling (GMP) layer is applied to gather important clues about distinctive object features. However, only a little information of high-dimensional features are effective \cite{DBLP:journals/cviu/0001EGB13}, e.g.clothes and shoes. Thus, a gated block (GB) is proposed to produce an attention map that focuses on meaningful features of both modalities. It selectively enforces informative features and mitigates the noise in the maxed-pooled features. Finally, we conduct extensive experiments on CUHK-PEDES, and our approach achieves the rank-1 score of 55.81\%. To our best knowledge, this is a new state-of-the-art.
	
	All in all, our contributions can be summarized as follows:
	\begin{itemize}
		\item The pre-trained language model, BERT, is innovatively integrated with CNN to learn better word embeddings for text-based person Re-id. Our experiments show that BERT can significantly improve the performance of the model and achieve better performance than other word embeddings.
		
		\item We propose a novel Dual-path CNN with Max Gated block (DCMG), which is based on two deep residual CNNs jointly optimized with CMPC loss and CMPM loss to embed the two modalities into a joint feature space. The GMP is applied to make visual-textual association concern more on remarkable features of both modalities for more reasonable similarity measurement. To further alleviate the noise of the maxed-pooled features, the gated block (GB) is proposed to produce an attention map that focuses on what is meaningful in the feature.
		
		\item Extensive ablation studies validate the effectiveness and complementarity of the proposed methods, and we obtain the state-of-the-art performance on the CUHK-PEDES dataset, outperforming other previous methods.
		
	\end{itemize}

	\section{Related work}
	
	In this section, we briefly introduce the related work about prior studies on word embedding, person re-identification, as well as text-based person re-identification.
	
	\subsection{Word Embedding}
	For natural language processing (NLP) \cite{wang2019atm,tang2019hidden,rong2019deep}, word embedding has become an essential component for various tasks. Word2vec \cite{DBLP:journals/corr/abs-1301-3781} can learn the context information of words to obtain word representations. Nevertheless, it only concerns the local information of the word and ignores the relevancy with the words outside the local window. Glove \cite{DBLP:conf/emnlp/PenningtonSM14} utilizes local and global information simultaneously. Different from word2vec and glove, EMLO \cite{DBLP:conf/naacl/PetersNIGCLZ18} dynamically adjusts word embedding and addresses the problem of polysemy with bi-LSTM. GPT \cite{radford2018improving} replaces bi-LSTM with Transformer \cite{DBLP:conf/nips/VaswaniSPUJGKP17}, which consists of the Multi-Head attention mechanism, to better capture long-distance language semantics. Moreover, BERT \cite{DBLP:conf/naacl/DevlinCLT19} leverages the bidirectional training of the Transformer to learn the contextual relations between the words in a textual description. It also introduces the word masking mechanism to learn the robust word embedding.
	
	In this paper, our method combines CNN with BERT to learn better word embeddings for text-based person Re-id and achieves better performance than other word embeddings.
	
	\subsection{Person Re-identification}
	Person re-identification (Re-id) has gained notable attention in all of its aspects. The early approaches focus on improving hand-crafted features and state-of-the-art approaches are mostly dominated by deep CNNs. The combination of global feature and partial feature is an essential solution to improve the performances in person Re-id task. Wang et.al \cite{DBLP:conf/mm/WangYCLZ18} proposed a multi-granularity network (MGN), which integrates discriminative information with various granularities. To utilize the spatial-temporal information, Wang et.al \cite{DBLP:conf/aaai/WangLHX19} constructed a two-stream spatial-temporal person Re-id framework to mine both visual semantic information and spatial-temporal information. Considering the pixel-level accuracy and capability of modeling arbitrary contours of human semantic parsing, Kalayeh et.al \cite{DBLP:conf/cvpr/KalayehBGKS18} integrated human semantic parsing in person Re-id. To obtain unique pose information, Sarfraz et.al \cite{DBLP:conf/cvpr/SarfrazSES18} incorporated both the fine and coarse pose information of the person to learn a discriminative embedding. Recently, the development of the generative adversarial network (GAN) \cite{DBLP:conf/nips/GoodfellowPMXWOCB14} also promotes the field of person Re-id, such as SPGAN \cite{DBLP:conf/cvpr/Deng0YK0J18}. AANet \cite{DBLP:conf/cvpr/TayRY19} is an attribute-based person Re-id method, which integrates person attributes and attribute attention maps into a classification framework.

	Unlike the traditional person Re-id approaches that exploit additional uni-modal clues, such as pose and mask, etc, text-based person Re-id is a multi-modal problem where the major challenge is the modality heterogeneity.
	
	\subsection{Text-Based Person Re-identification}
	In the initial work, the first large scale person Re-id data (CUHK-PEDES), which contains person images with detailed natural language annotations, was introduced by Li et al \cite{DBLP:conf/cvpr/LiXLZYW17}. They also proposed the Recurrent Neural Network with Gated Neural Attention mechanism model (GNA-RNN) to learn the word by word affinity of a sentence with image features. To effectively utilize identity-level annotations, Li et al \cite{DBLP:conf/iccv/LiXLYW17} further used a two-stage CNN-LSTM network for textual-visual matching. To alleviate the cross-modal fine-grained problem, Niu et al. \cite{DBLP:journals/tip/NiuHOW20} achieved better cross-modal similarity evaluation by a Multi-granularity Image-text Alignments (MIA) model. Recently, Nikolaos \cite{DBLP:conf/iccv/Sarafianos0K19} combined BERT with bi-LSTM to learn better word embedding and introduced a Text-Image Modality Adversarial Matching (TIMAM) approach, which learns modality-invariant feature representations with adversarial and cross-modal matching objectives. Different from the above methods which are all CNN-RNN architectures, Zheng et al \cite{zheng2020dual}, proposed a dual CNN architecture to extract image-text features and used each image-sentence pair as a single class to train the whole network with instance loss and ranking loss.
	
	Beyond the aforementioned methods, we are probably the first to combine BERT with CNN to extract word embeddings for text-based person Re-id. We further utilize GMP and gated block to make visual-textual association concern more on remarkable features of both modalities for more reasonable similarity measurement.

	\begin{figure*}
		\begin{center}
			\includegraphics[width=1\linewidth]{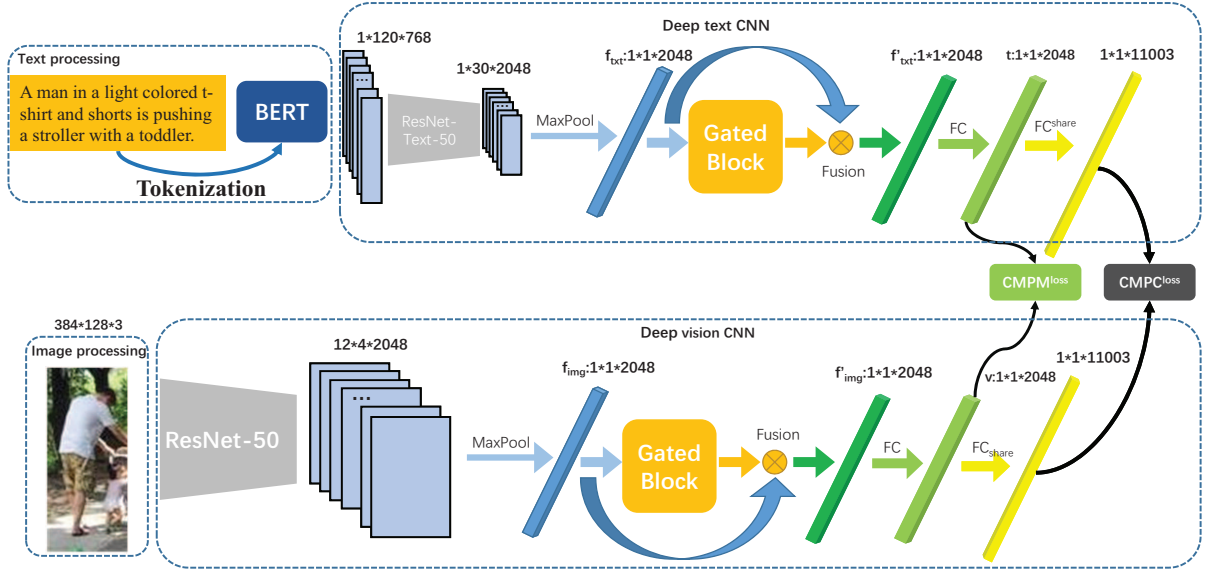} \ \\
		\end{center}
		\caption{DCMG is based on two deep residual CNNs jointly optimized with CMPC loss and CMPM loss to embed the two modalities into a joint feature space, i.e., deep text CNN (top) and deep vision CNN (bottom). The deep vision CNN is based on ResNet-50 pre-trained on ImageNet. The deep text CNN is similar to the vision CNN but the convolution filters are of size $ 1\times 3 $ instead of $ 3\times 3 $. The GB is composed of a multi-layer perception with two hidden layers (MLP) and produces an attention map that focuses on what is meaningful in the maxed-pooled features. After the GB, we add one fully connected layer in both vision CNN and text CNN to obtain the final image descriptor $ v $ and text descriptor $ t $. Finally, we add a shared-weight $ {{W}_{share}} $ classification layer}
		\label{fig:DCMG}
	\end{figure*}
	\begin{figure*}
		\begin{center}
			\includegraphics[width=1\linewidth]{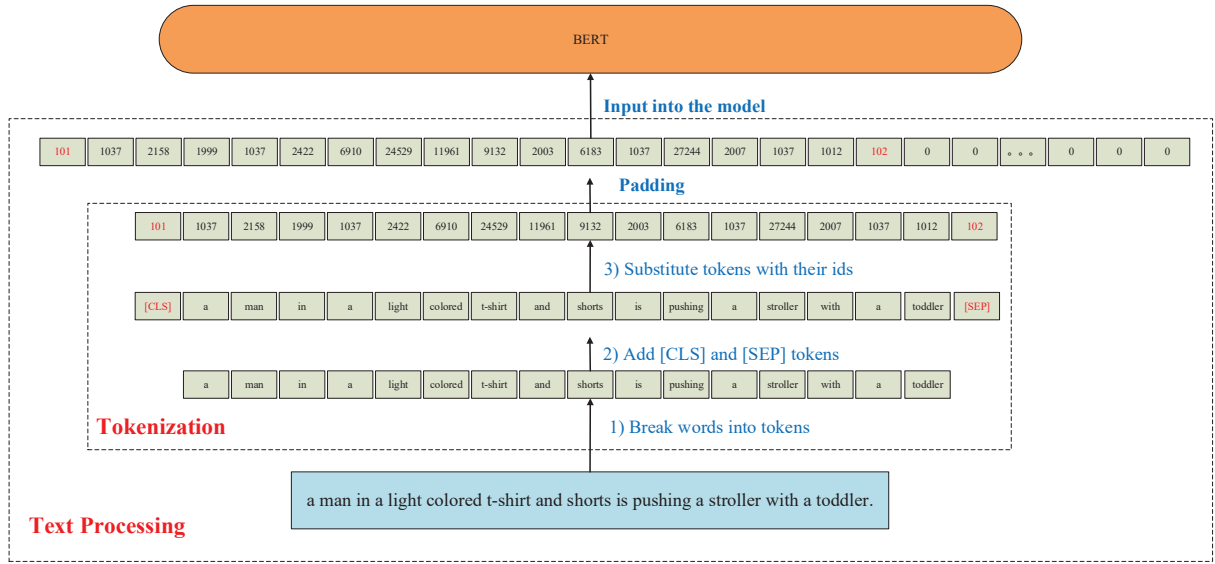} \ \\
		\end{center}
		\caption{Some minimal processing is needed to do to put the sentence in the format BERT requires. Our first step is to tokenize the sentences, which breaks them up into words and subwords in the format BERT is comfortable with. We then replace the tokens with corresponding numbers in the embedded table, which can be got from the pre-training model. After tokenization, each sentence is represented as a list of tokens. To process the examples all at once, we need to pad all lists to the same size.}
		\label{fig:token}
	\end{figure*}

	\section{The Proposed Methodology}
	In this section, the novel Dual-path CNN with Max Gated block (DCMG) is introduced, which is jointly optimized with 
	cross-modal projection matching (CMPM) loss and cross-modal projection classification (CMPC) loss to embed the two modalities into a joint feature space, as presented in Figure \ref{fig:DCMG}. We use BERT \cite{DBLP:conf/naacl/DevlinCLT19} for natural language representation. Before the sentence is handed to BERT, text processing is needed to put the sentence in the format BERT requires. Then, each sentence can be converted to word embeddings. Data augmentation techniques are also applied in image processing to create diversity in the training dataset.
	There are dual-path CNN inside the framework: (1) deep text CNN and (2) deep vision CNN. In deep text CNN, ResNet-Text-50, which is a variant of ResNet-50, is proposed to extract the textual feature maps. In deep vision CNN, the ResNet-50, which is pre-trained on ImageNet, extracts the visual feature maps. We use global max pooling (GMP) to obtain global information on textual and visual feature maps. To further alleviate the noise of the maxed-pooled features, the gated block (GB) is proposed to produce an attention map that focuses on meaningful features of both modalities. Then, we add one fully connected (FC) layer in both vision CNN and text CNN to obtain the final image descriptor and text descriptor, which are used to perform text-to-image retrieval. Finally, we add a shared-weight classification layer. By combining two CNNs, we can obtain discriminative visual embeddings for cross-modal similarity evaluation.
	
	\subsection{Text and Image Processing}
	\textbf{Text processing:}To extract better word embeddings, we apply a recently proposed language representation model named BERT. Before we input sentence to BERT, some minimal processing is needed to turn them into the format BERT requires, shown in Figure \ref{fig:token}. Our first step is to tokenize the sentences, which breaks them up into words and subwords in the format BERT is comfortable with. Specifically, we add the special words to the sentence (add [CLS] at the bedding of the sentence and [SEP] at the end). We then replace the tokens with corresponding numbers in the embedded table, which can be obtained from the pre-training model. After tokenization, each sentence is represented as a list of tokens. To process the examples all at once, we need to pad all lists to the same size. For the integrity of the description, the fixed-length of the sentence is set to 120 words, which is the max length of samples. Our experiments show the effectiveness of long sentences. The sentences shorter than the fixed-length are zero-padded. 
	
	After that, each tokenized sentence is inputted to the BERT to extract a 768-D word embedding for each word. Thus, each sentence can be converted to word embeddings of size $ 120\times 768 $. We further reshape the word embeddings into the $ 1\times 120\times 768 $ format, which can be considered as height, width, and channel of images, respectively.
	
	\textbf{Image Processing: }Since pedestrians are usually rectangular, all images are resized to 384x128. For data augmentation, we pad images on all sides with 10 values and random crops of the original dimensions are extracted.To make the model more robust, we horizontally flip the images randomly with 50\% probability and shuffle the images. Finally, the images are normalized with mean and standard deviation.

	\subsection{Network Architecture}
	{\bf Deep text CNN}:In Figure \ref{fig:DCMG}, the upper branch corresponds to deep text CNN, which is a 50 layered deep ResNet modified to deal with word embeddings. The filter size of the first convolutional layer is $ 1\times 1\times 300 $ and the max-pooling layer after the first convolutional layer is removed. The rest of the network is similar to the ResNet-50 except that the convolution filters are of size $ 1\times 3 $ instead of $ 3\times 3 $, which produce responses by looking at the three neighboring words. The $ 1\times 30\times 2048 $ feature map $ {{T}_{txt}} $ can be extracted after the word embeddings are inputted to the text CNN.

	{\bf Deep vision CNN}:
	In Figure \ref{fig:DCMG}, the lower branch is vision CNN which is based on ResNet-50 pre-trained on ImageNet. The structure before the original global average-pooling layer (GAP) is maintained the same as the original architecture. The GAP layer and what follows are removed. Given an input image of size $ 384\times 128 $, a forward pass of the network produces a $12\times 4\times2048  $ feature map $ {{T}_{img}} $ of activation.
	
	The visual-textual feature maps $ {{T}_{txt}} $ and ${{T}_{img}} $ are obtained by the vision CNN and the text CNN, respectively. To make the visual-textual features focus more on discriminative characteristics, i.e. salient part, we use the global Max-pooling (GMP) layer to gather important clues about distinctive object features. Then, we can obtain two 2048-dimensional feature vectors, $ {{f}_{img}} $ and $ {{f}_{txt}} $.

	
	\begin{figure*}
		\begin{center}
			\includegraphics[width=0.5\linewidth]{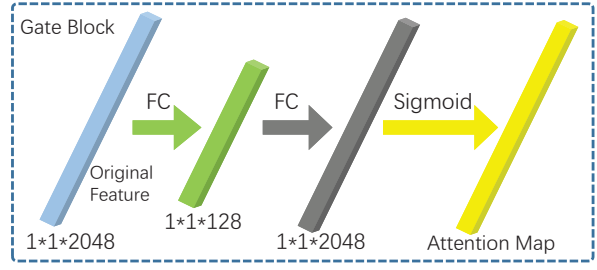} \ \\
		\end{center}
		\caption{Gated block}
		\label{fig:GB}
	\end{figure*}

	{\bf Gated block (GB)}:
	The GMP activates on more local saliency, but only a few features of high-dimensional features are effective and max pooling will introduce a lot of noise. Thus,  GB is proposed to produce an attention map to selectively enforce informative features and suppress the noise in the maxed-pooled features, as shown in Figure \ref{fig:GB}. The GB is composed of a multi-layer perception with two hidden layers (MLP), which is defined as:
	
	\begin{equation}\label{key}
	{{F}^{gate}}\left( f \right)=\sigma \left( {{W}_{2}}\delta \left( {{W}_{1}}f \right) \right),
	\end{equation}
	where $ \delta  $ denotes the ReLU function, $ {{W}_{1}}\in {{\mathbb{R}}^{\frac{2048}{r}\times 2048}} $ and $ {{W}_{2}}\in {{\mathbb{R}}^{2048\times \frac{2048}{r}}} $. To limit model complexity and add generalization, the first hidden layer is a dimensionality-reduction layer with parameters $ {{W}_{1}} $ and reduction radio $ r $. Then feature vectors, $ {{f}_{img}} $  and $ {{f}_{txt}} $, are rescaled with the attention map: 
	\begin{equation}\label{key}
	{{{f}'}_{img}}={{f}_{img}}\otimes {{F}^{gate}}\left( {{f}_{img}} \right)
	\end{equation}
	\begin{equation}\label{key}
	{{{f}'}_{txt}}={{f}_{txt}}\otimes {{F}^{gate}}\left( {{f}_{txt}} \right)
	\end{equation}
	
	The GB introduces dynamics conditioned on the features, helping to boost feature discriminability. The final image descriptor, $ v $, and text descriptor, $ t $, are obtained after a fully-connected layer (input dim:$ 2048 $, output dim:$ 2048 $). We also add a shared-weight $ {{W}_{share}} $ classification layer (input dim:$ 2048 $, output dim: $ 11003 $).

	\subsection{Objective Function}
	We employ two kinds of loss functions, i,e., identification loss and matching loss that bring the features originated from the same identity close and push away features originated from different identities.
	
	{\bf Identification loss}:
	We regard different IDs in the training set as the number of categories and classify images and descriptions to the corresponding ID category separately. The identification loss is CMPC loss where the cross-modal projection is integrated into the norm-softmax loss to further enhance the compactness of the matched embeddings. 
	
	Given a mini-batch with $ N $ image-text pairs, image descriptors, $ V=\left\{ {{v}_{i}} \right\}_{i=1}^{N} $, and text descriptors, $ T=\left\{ {{t}_{i}} \right\}_{i=1}^{N} $ can be extracted. The label set $ Y=\left\{ {{y}_{i}} \right\}_{i=1}^{N} $ is from $ M $ classes, then the CMPC loss for classifying image can be described as follows:
	\begin{equation}\label{key}
	L_{cmpc}^{ipt}=\frac{1}{N}\sum\limits_{i=1}^{N}{\log \left( \frac{\exp \left( W_{{{y}_{i}}}^{T}{{{\hat{v}}}_{i}} \right)}{\sum\limits_{j=1}^{M}{\exp \left( W_{j}^{T}{{{\hat{v}}}_{i}} \right)}} \right)}
	\end{equation}
	
	\begin{equation}\label{key}
	\left\| {{W_j}} \right\| = 1,{{\hat v}_i} = v_i^T{{\bar t}_i} \cdot {{\bar t}_i},{{\bar t}_i} = \frac{{{t_i}}}{{\left\| {{t_i}} \right\|}},
	\end{equation}
	where $ {{\bar{t}}_{i}} $ denotes the normalized text feature, $ {{\hat{v}}_{i}} $ represents the vector projection of image feature $ {{v}_{i}} $ onto the normalized text feature $ {{\bar{t}}_{i}} $, $ N $ is the batch size, $ {{W}_{j}} $ and $ {{W}_{{{y}_{i}}}} $ are the $ {{y}_{i}}-th $ and $ j-th $ column of the weight matrix $ W $ in the classification layer. The text classification loss function is computed in a similar manner and the final CMPC loss can be calculated with:
	\begin{equation}\label{key}
	{{L}_{cmpc}}=L_{cmpc}^{ipt}+L_{cmpc}^{tpi}
	\end{equation}
	
	{\bf Matching loss}:
	To refrain from using loss that sample triplets or quadruplets within each batch, which introduce a computational overhead during training to obtain hard negatives, we use CMPM loss which incorporates the cross-modal projection into KL divergence to associate the representations across different modalities.
	
	Given $ n $ image and text samples in a batch, for each image feature, $ {{v}_{i}} $, the image-text feature pairs are constructed as $ \left\{ \left( {{v}_{i}},{{t}_{j}} \right),{{m}_{i,j}} \right\}_{j=1}^{n} $, where $ {{m}_{i,j}}=1 $ means that $ {{v}_{i}} $ and $ {{t}_{j}} $ are the same ID category, while $ {{m}_{i,j}}=0 $ indicates the unmatched ones. The probability of matching $ {{v}_{i}} $ to $ {{t}_{j}} $ is defined as:
	\begin{equation}\label{key}
	{{p}_{i,j}}=\frac{\exp \left( v_{i}^{T}{{{\bar{t}}}_{j}} \right)}{\sum\limits_{k=1}^{n}{\exp \left( v_{i}^{T}{{{\bar{t}}}_{k}} \right)}}
	\end{equation}
	\begin{equation}\label{key}
	{{\bar{t}}_{j}}=\frac{{{t}_{i}}}{\left\| {{t}_{i}} \right\|},
	\end{equation}
	where $ {{p}_{i,j}} $ denotes the proportion of scalar projection of $ \left( {{v}_{i}},{{t}_{j}} \right) $ among all pairs $ \left\{ \left( {{v}_{i}},{{t}_{j}} \right) \right\}_{j=1}^{n} $ in a batch. Thus, the more similar the image feature to text feature, the larger the scalar projection is from former to the latter. Consider the fact that in each batch there might be more than one matched text samples for image sample $ {{v}_{i}} $, the true matching probability is normalized as: 
	\begin{equation}\label{key}
	{{q}_{i,j}}=\frac{{{m}_{i,j}}}{\sum\limits_{k=1}^{n}{{{m}_{i,k}}}}
	\end{equation}
	
	The CMPM loss of associating $ {{v}_{i}} $ with correctly matched text samples for each bach is defined as:
	\begin{equation}\label{key}
	L_{cmpm}^{ipt}=-\frac{1}{n}\sum\limits_{i=1}^{n}{\sum\limits_{j=1}^{n}{{{p}_{i,j}}\log \left( \frac{{{p}_{i,j}}}{{{q}_{i,j}}+\varepsilon } \right)}},
	\end{equation}
	where $ \varepsilon  $ is a small number for preventing numerical problems. The same procedure is followed to perform the opposite matching (\textit{i.e.,} from text to image) to compute loss $ L_{cmpm}^{tpi} $ and the bi-directional CMPM loss is computed by:
	\begin{equation}\label{key}
	{{L}_{cmpm}}=L_{cmpm}^{ipt}+L_{cmpm}^{tpi}
	\end{equation}
	
	The overall objective function is formulated as:
	\begin{equation}\label{key}
	L={{L}_{cmpc}}+{{L}_{cmpm}}
	\end{equation}

	\section{Experiments and Analysis}

	\subsection{Dataset}
	
	The CUHK-PEDES is the only large scale available dataset for the text-based person Re-id \cite{DBLP:conf/cvpr/LiXLZYW17}. It contains 40206 pedestrian images and 80412 textual descriptions by combining several Re-id datasets, for 13003 identities. Each image is annotated by two textual descriptions. On average, each person has 3.1 images and each textual description contains more than 23 words. The dataset contains 9408 different words. The data split method proposed by Li et al \cite{DBLP:conf/cvpr/LiXLZYW17} is utilized in our experiments. The training set has 34054 images, 11003 persons, and 68126 textual descriptions. The validation set has 3078 images, 1000 persons, and 6158 textual descriptions. The test set has 3074 images, 1000 persons, and 6156 textual descriptions.

	\subsection{Configuration}
	\textbf{Implementation details:} We use PyTorch as our deep learning framework and a single NVIDIA GeForce GTX 1080ti GPU. The model is trained for 160 epochs using stochastic gradient descent (SGD) algorithm with momentum fixed to 0.9. The size of the image feature $ v $  and text feature $ t $  is equal to 2048. The reduction ratio $ r $  is 16. The batch size is 64 and the initial learning rate is set to 0.1. The warm-up learning rate as descripted in equal \ref{lr} is adopted to make the model stable and converge fast.
	
	The learning rate $ lr\left( e \right) $ in epoch $ e $ is computed as:
	\begin{equation}\label{lr}
	lr\left( e \right)=\left\{ \begin{aligned}
	& e*\frac{0.1}{10},e\le 10 \\ 
	& 0.1,10<e\le 55 \\ 
	& 0.01,55<e\le 80 \\ 
	& 0.001,80<e\le 100 \\ 
	& 0.0001,100<e\le 120 \\ 
	& 0.00001,120<e\le 140 \\ 
	\end{aligned} \right.	
	\end{equation}
	
	We split the model training into two steps. We fix the weights of BERT in the whole training process. In step-1, we fix the parameters of pre-trained visual CNN and only train the other model parameters for 80 epochs. In stage-2, we release the parts of the visual CNN and train the visual CNN and text CNN for 80 epochs.
	
	\textbf{Evaluation metrics:} At testing time, features $ f $ (i.e. image descriptor $ v $ and text descriptor $ t $) are extracted through the network. The distance between the query (text) and the gallery (image) is computed using the cosine similarity:
	\begin{equation}\label{key}
	S\left( {{f}_{query}},{{f}_{gallery}} \right)=\frac{{{f}_{query}}\centerdot {{f}_{gallery}}}{\left\| {{f}_{query}} \right\|\centerdot \left\| {{f}_{gallery}} \right\|}
	\end{equation}
	
	The distances are then sorted. We adopt rank-1, rank-5, and rank-10 accuracies to evaluate the performance.  Given a textual description, we rank all test images by their similarities with the query text. If top-K images contain any corresponding person, the search is successful.

	\subsection{Results on the CUHK-PEDES}
	
	\begin{table}[htb]
		\caption{Results (\%) on the CUHK-PEDES dataset. Results are ordered based on the rank-1 accuracy.}
		\small 
		\label{tab1}
		\centering
		\begin{tabular}{l rcc}
			\toprule
			Method         & Rank-1         & Rank-5         & Rank-10        \\ \midrule
			GNA-RNN~\cite{DBLP:conf/cvpr/LiXLZYW17}        & 19.05          &                & 53.64          \\ 
			IATVM~\cite{DBLP:conf/iccv/LiXLYW17}          & 25.94          &                & 60.48          \\ 
			PWM-ATH~\cite{DBLP:conf/wacv/ChenXL18}        & 27.14          & 49.45          & 61.02          \\  
			Dual-Path~\cite{zheng2020dual}      & 44.40          & 66.26          & 75.07          \\  
			CMPM+CMPC~\cite{DBLP:conf/eccv/ZhangL18}      & 49.37          &                & 79.27          \\ 
			MCCL~\cite{DBLP:conf/icassp/WangBWWQL19}           & 50.58          &                & 79.06          \\ 
			MIA~\cite{DBLP:journals/tip/NiuHOW20}            & 53.10          & 75.00          & 82.90          \\ 
			PMA~\cite{DBLP:conf/aaai/JingSW00T20}            & 53.81          & 73.54          & 81.23          \\ 
			TIMAM~\cite{DBLP:conf/iccv/Sarafianos0K19}          & 54.51          & \textbf{77.56}          & 84.78          \\ \midrule
			\textbf{DCMG} & \textbf{55.81} & 77.44 & \textbf{84.87} \\ \bottomrule
			
		\end{tabular}
		\vspace{-0.35cm}
	\end{table}

	Table 1 shows the results of our proposed DCMG and the nine best-performance methods on the CUHK-PEDES. The nine methods that have been evaluated on this dataset include (1) GNA-RNN \cite{DBLP:conf/cvpr/LiXLZYW17} which applies unit-level attentions and word-level gates to learn the word by word affinity of a sentence with image features. (2) IATVM \cite{DBLP:conf/iccv/LiXLYW17} which is a two-stage CNN-LSTM network with two attention modules working on two modalities at different levels. (3) PWM-ATH \cite{DBLP:conf/wacv/ChenXL18} which utilizes a patch-word matching and learns an adaptive threshold for each word to enhance or suppress its effect on the final text-image affinity. (4) Dual-path \cite{zheng2020dual}  which has similar dual-path CNN architecture to extract image-text features optimized with ranking loss and instance loss. (5) CMPM+CMPC \cite{DBLP:conf/eccv/ZhangL18}  which proposes CMPM loss and CMPC loss for learning deep discriminative image-text embeddings. (6) MCCL \cite{DBLP:conf/icassp/WangBWWQL19}  which propose a novel mutually connected classification loss to fully exploit the identity-level information. (7) MIA \cite{DBLP:journals/tip/NiuHOW20}  which combines three different granularities hierarchically, i.e., global-global, global-local, and local-local, to alleviate the cross-modal fine-grained problem. (8) PMA \cite{DBLP:conf/aaai/JingSW00T20}  which exploits the multilevel corresponding visual contents and learns latent semantic alignment between the visual body part and textual noun phrase. (9) TIMAM \cite{DBLP:conf/iccv/Sarafianos0K19} which combines BERT with bi-LSTM to learn better word embedding, and use adversarial and cross-modal matching objectives to learn modality-invariant feature representations.
	We observe that DCMG obtains the rank-1 score of 55.81\% and outperforms all compared methods in terms of rank-1, which demonstrate the effectiveness of our methods on the dataset. Although TIMAM obtains the highest accuracy in terms of rank-5, our approach outperforms TIMAM by 1.30\% in terms of rank-1 and 0.09\% in terms of rank-10, which are more important.

	\subsection{Qualitative Results}
	\begin{figure*}
		\begin{center}
			\includegraphics[width=0.8\linewidth]{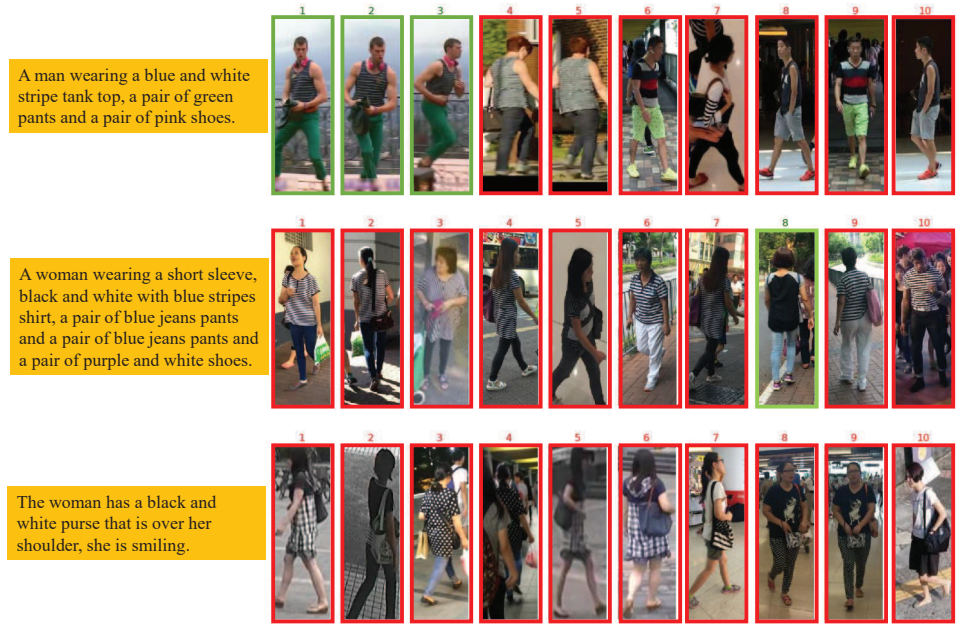} \ \\
		\end{center}
		\caption{Qualitative image search results using text query by our proposed DCMG. The results are sorted from left to right according to their confidence. The images in green boxes are the true matches, and the images in red boxes are the false matches. Successful searches (Rows 1-2)  where corresponding persons are in the top-10 results. Failure cases (Row 3) where corresponding persons are not in the top-10 results.}
		\label{fig:visual}
	\end{figure*}

	\begin{figure*}
		\begin{center}
			\includegraphics[width=0.8\linewidth]{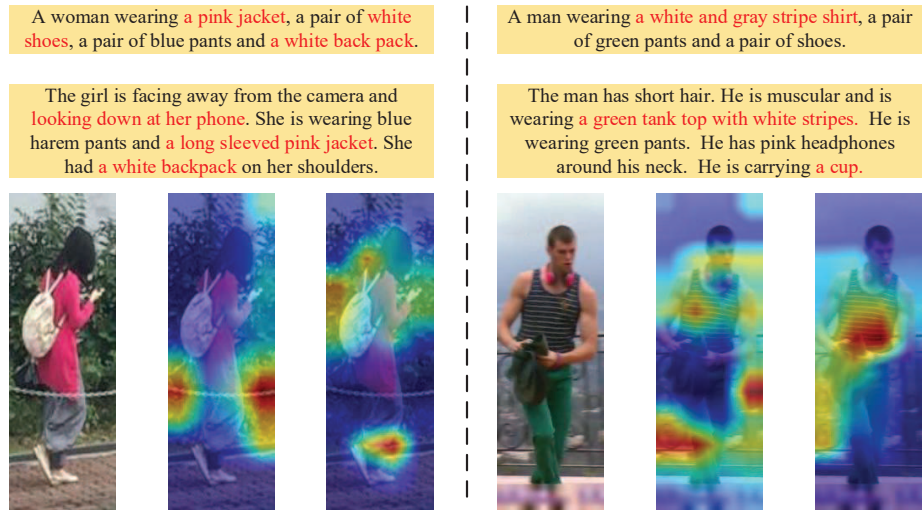} \ \\
		\end{center}
		\caption{The top 2 rows show the textual descriptions and the bottom of the row is triplet images, which contain, from left to right, original image, activation map of Pre-trained ResNet-50, and activation map of DCMG. These images and descriptions show that DCMG concerns more on the remarkable features of both modalities.}
		\label{fig:hotmap}
	\end{figure*}
	
	We conduct the qualitative evaluation on CUHK-PEDES for our proposed DCMG. Figure \ref{fig:visual} presents three visual retrieval results with natural language descriptions. The top 2 rows show successful cases where corresponding images within the top-10 retrieval results. For the successful cases, we can observe that the top results have multiple regions that fit the noun phrase in the text. Some non-corresponding images also show correlations to the query descriptions (e.g., a blue and white stripe tank top and a pair of green pants). Some interesting observations can be made from the failure cases. While all retrieved images in the bottom case have a different ID with the ground truth of the language description, the retrieved persons are similar to the textual descriptions, who all wear a black and white purse. It indicates that DCMG can successfully learn cloth and accessory-related things and retrieve images that match the textual input.
	
	To understand how DCMG learn discriminative visual-textual feature embeddings, we visualize the activations of the last convolutional feature maps to investigate where the network focuses on to extract features. Following \cite{DBLP:conf/iclr/ZagoruykoK17}, the activation maps are computed as the sum of absolute-valued feature maps along the channel dimension followed by a spatial l2 normalization. Figure \ref{fig:hotmap} compares the activation maps of DCMG and the Pre-trained ResNet-50. It demonstrates that DCMG can capture the local discriminative regions (e.g., white shoes and a white backpack), which match the corresponding person image to a textual description. In contrast, the Pre-trained ResNet-50 pays more attention to the uncorrelated regions. Therefore, this qualitative result shows that DCMG makes the visual-textual features focus more on discriminative characteristics, i.e. salient part, and gathers important clues about distinctive object features.
	
	\begin{table}[htb]
		
		\caption{Ablation studies on the CUHK-PEDES dataset to investigate the additions of DCMG in terms of rank-1, rank-5, and rank-10 accuracy for GAP, GMP, and GB.}
		\small 
		\centering
		\label{tab3}
		\begin{tabular}{ccc|ccc}\toprule
			GAP & GMP & GB & Rank1 & Rank5 & Rank10 \\ \midrule
			\checkmark	&      &             & 54.26 & 75.93 & 83.80  \\
			\checkmark	&     &   \checkmark          & 54.77 & 76.51 & 84.37  \\
			& \checkmark    &             & 54.26 & 76.29 & 84.06  \\
			& \checkmark    &\checkmark             & \textbf{55.81} & \textbf{77.44} &\textbf{84.87 } \\
			\checkmark&\checkmark     &             & 54.28 & 76.07 & 83.93  \\
			\checkmark&     \checkmark&\checkmark             & 54.98 & 77.44 & 84.68 \\ \bottomrule
		\end{tabular}
		\vspace{-0.35cm}
	\end{table}
	
	\begin{table}[htb]
		
		\small 
		\caption{Ablation studies on the CUHK-PEDES datasets to investigate the CMPM loss and CMPC loss in terms of rank-1, rank-5, and rank-10 accuracy.}
		
		\centering
		\label{tab4}
		\begin{tabular}{cc|ccc}
			\toprule
			CMPM & CMPC & Rank-1 & Rank-5 & Rank-10 \\ \midrule
			\checkmark    &      & 54.20     & 75.99     & 84.00     \\
			& \checkmark    & 29.26     & 51.06     & 60.77     \\
			\checkmark    & \checkmark    & \textbf{55.81}     & \textbf{77.44}     & \textbf{84.87}    \\ \bottomrule
		\end{tabular}
		\vspace{-0.35cm}
	\end{table}
	
	\begin{table}[htb]
		
		\small 
		\caption{Ablation studies on the CUHK-PEDES dataset to assess the impact of the fixed- length of descriptions.}
		\centering
		\label{tab5}
		\begin{tabular}{l|ccc}
			\toprule
			Length & Rank-1 & Rank-5 & Rank-10 \\ \midrule
			40     & 52.93  & 76.12  & 84.08   \\
			60     & 53.97  & 76.85  & 84.78   \\
			80     & 54.44  & 77.31  & 85.28   \\
			100    & 54.64  & 77.00  & 84.37   \\
			120     & \textbf{55.81}     & \textbf{77.44}     & \textbf{84.87}   \\ \bottomrule
		\end{tabular}
		\vspace{-0.35cm}
	\end{table}
	
	\begin{table}[htb]
		
		\small 
		\caption{Experiments on the CUHK-PEDES datasets to compare several word embeddings with BERT in terms of rank-1, rank-5, and rank-10 accuracy.}
		\centering
		\label{tab6}
		\begin{tabular}{l|lll}
			\toprule
			Word Embedding & Rank-1 & Rank-5 & Rank-10 \\ \midrule
			Glove50        & 52.13       & 74.48       & 82.55        \\
			Glove100       &  52.70      & 74.90       &  82.47       \\
			Glove200       & 52.46      & 74.59       &82.84         \\
			Glove300       &  52.77      & 75.37       & 82.82        \\
			BERT           & \textbf{55.81}     & \textbf{77.44}     & \textbf{84.87}   \\ \bottomrule
		\end{tabular}
		\vspace{-0.35cm}
	\end{table}

	\begin{table}[]
		\small 
		
		\caption{Experiments on the CUHK-PEDES datasets to assess the training strategies and the pre-trained visual CNN in terms of rank-1, rank-5, and rank-10 accuracy.}
		\centering
		\label{tab7}
		\setlength{\tabcolsep}{1mm}{
			\begin{tabular}{cccc|ccc}
				\toprule
				Strategy & Pre-trained & Stage1 & Stage2 & Rank1 & Rank5 & Rank10 \\ \midrule
				1        &             &        & \checkmark       & 53.27 & 74.85 & 82.91  \\
				2        &\checkmark             &\checkmark        &        & 16.75 & 36.47 & 47.43  \\
				3        &\checkmark             &        &\checkmark        & 54.10 & 76.22 & 83.72  \\
				4        &\checkmark             &\checkmark        &\checkmark        & \textbf{55.81}     & \textbf{77.44}     & \textbf{84.87}  \\ \bottomrule
			\end{tabular}
		}
		\vspace{-0.35cm}
	\end{table}

	\subsection{Ablation study}
	To investigate the several components in DCMG, we perform a set of ablation studies and assess how each component of DCMG contributes to the final matching performance on the CUHK-PEDES dataset. 
	
	{\bf Impact of proposed components}:
	In our first ablation study, the influence of GAP, GMP, and GB on the CUHK-PEDES is investigated in Table \ref{tab3}. We first test only applying pooling operation in the network. As shown by the results, the concatenating average-pooled and max-pooled features have the best results in terms of rank-1, which both contain the information of max-pooled and average-pooled feature maps. However, in terms of rank-5 and rank-10, GMP shows the best performance.

	In the following experiments, We combine GB with different pooling layers. The results show that GB can improve the performance of all pooling operations, which demonstrates that GB can selectively enforce informative features and suppress the noise in the features. Compared with others, GMP+GB is the best match, which can better utilize the relationship between the image area and the noun phrase in the text than Average Pooling.

	{\bf Loss functions}:
	In the second ablation study, Table \ref{tab4}  illustrates the impact of the CMPM loss and CMPC loss in image-text embedding learning on the CUHK-PEDES dataset. We notice that the CMPM loss alone achieves competitive results. While the performance of CMPC loss is common, the rank-1 rate of CMPM loss is increased from 54.20\% to 55.71\%, via the CMPC loss. The results verify that CMPC loss further improves the performance of CMPM loss and two losses can work together to improve the final retrieval results. On the one hand, CMPM loss increases the variance between unmatched samples and the association between the matched ones. On the other hand, CMPC loss does help to regularize the model.

	{\bf Impact of description length}:
	We then investigate the influence of different description lengths. As shown in Table \ref{tab5}, DCMG yields a steady improvement throughout the increasing of description length and DCMG with fixed-length of 120 achieves the best results. We argue that the long description can ensure the integrity of the description and avoid the risking of losing critical clues, because 120 is the max length of samples. What is more, CNN overcomes the limitation of parallelization at longer sequence lengths, thus, the additional computational overhead is acceptable.

	{\bf Discussion of word embedding}:
	In this section, we compare several word embeddings with BERT, i,e., glove50, glove100, glove200, and glove300, which extract a 50-D, 100-D, 200-D, and 300-D word embedding for each word, respectively. Note that we remove the words, which have not appeared in the training set as well as dictionary. For CUHK-PEDES, the dictionary size is 7012. When testing, the missing words in the dictionary will also be removed in advance. Glove initializes the embedding layer and then generates word embedding. Meanwhile, the embedding layer is fixed during the whole training step.
	
	As shown in Table \ref{tab6}, different types of glove word embeddings have similar performance in terms of rank-1. The rank-1 rate of BETR outperforms the glove300 by 2.94\%, which the best result of glove word embeddings. The results indicate that BERT can successfully be combined with text CNN and significantly improve the performance of the model, which originates from that BERT learns the contextual relations between the words in a textual description and introduces a word masking mechanism to learn robust word embedding.

	{\bf Discussion of Training Strategies}:
	To avoid compromising one modality with another and gain better insight into the joint training of a multi-modal system, the retrieval performance of four training strategies defined by various initialization of the weights and learning schemes has been analyzed. For text CNN, we initialize it with Xavier initialization \cite{glorot2010understanding}. For vision CNN, we initialize it with a pre-trained ImageNet model, ResNet-50, except from strategy 1, which initializes vision CNN with Xavier initialization. Specifically, strategy 1 and strategy 3 train the two CNNs simultaneously. Strategy 2 fixes the vision CNN and updates the parameters of the text CNN throughout the training process. Strategy 4 fixes the parameters of pre-trained visual CNN and only trains the other model parameters on stage 1. Then, the whole network is trained until it converges on stage 2.
	
	Table \ref{tab7} compare the performance of different training strategies. Compared with strategy 2, strategy 3 with ImageNet initialization increases the rate of rank-1 from 53.27\% to 54.10\%. It indicates that ImageNet initialization offers a better starting point and a wider learning surface as compared to the visual CNN training from scratch. Thus, the choice of initial weights is greatly affecting the results. For strategy 2, freezing the parameters of visual CNN and only tuning the parameters of the text CNN throughout the training phase will greatly reduce the accuracy of the retrieval results. On the contrary, strategy 4 achieves the best results, which fixes the vision CNN in the first stage and tunes the vision and text CNN simultaneously in the second stage, due to the fact that it may be hard to train the two CNNs simultaneously at the beginning. The results suggest that our two stage  strategy is most suitable to be adopted in the experiments.

	\section{Conclusions}
	
	To make the visual-textual feature focus more on discriminative characteristics, we propose a novel dual-path CNN with max gated block (DCMG) for text-based person Re-id in this paper. The global Max pooling (GMP) layer is applied to makes the visual-textual feature focus more on discriminative characteristics, i.e. salient part, and the GB produces an attention map to selectively enforce informative features and suppress the noise in the maxed-pooled features. To the best of our knowledge, we are probably the first to combine the pre-trained language model, BERT, with CNN to learn better word embedding for text-based person Re-id. Extensive experiments with ablation analysis validate the effectiveness and complementarity of the proposed methods and our approach obtains the state-of-the-art performance on the CUHK-PEDES dataset.
	

	\section{Acknowledgement}
	This work was supported in part by National Science Foundation of China ( No.U1736105). The authors extend their appreciation to the Deanship of Scientific Research at King Saud University for funding this work through research group no. RG-1441-331.

	\bibliographystyle{IEEEtran}
	\bibliography{latexfile}

	%

	
	

\end{document}